\documentclass[tablecaption=bottom,wcp]{jmlr} 
\usepackage[utf8]{inputenc}

\usepackage{longtable}

\usepackage{booktabs}

\theorembodyfont{\upshape}
\theoremheaderfont{\scshape}
\theorempostheader{:}
\theoremsep{\newline}

\usepackage{caption}
\usepackage{subcaption}
\usepackage{listings, chngcntr, float, amsmath, amssymb, cmap, graphicx, xcolor, hyperref}
\usepackage{amsmath, multirow}

\usepackage{color}
\usepackage{grffile}
\usepackage{array,tabulary}
\usepackage{wrapfig}
\graphicspath{{media/}}

\usepackage{mathtools,mathptmx}
\mathtoolsset{showonlyrefs=true}

\usepackage[mathcal]{euscript}
\usepackage{amsfonts}
\usepackage{tikz}
\usetikzlibrary{arrows.meta,shapes.misc}

\jmlrvolume{60}
\jmlryear{2018}
\jmlrworkshop{Conformal and Probabilistic Prediction and Applications}
\jmlrproceedings{PMLR}{Proceedings of Machine Learning Research}

\title[ML pipeline for discovering neuroimaging-based biomarkers]{Machine Learning pipeline for discovering neuroimaging-based biomarkers in neurology and psychiatry}

\newcommand*{\mathcolor}{}
\def\mathcolor#1#{\mathcoloraux{#1}}
\newcommand*{\mathcoloraux}[3]{%
  \protect\leavevmode
  \begingroup
    \color#1{#2}#3%
  \endgroup
}

\author{\Name{Alexander Bernstein} \Email{a.bernstein@skoltech.ru}\\
\addr Skolkovo Institute of Science and Technology,  Skolkovo, Moscow Region, Russia
\AND
\Name{Evgeny Burnaev} \Email{e.burnaev@skoltech.ru}\\
\addr Skolkovo Institute of Science and Technology,  Skolkovo, Moscow Region, Russia
\AND
\Name{Ekaterina Kondratyeva}\Email{ekaterina.kondratyeva@skolkovotech.ru}\\
\addr Skolkovo Institute of Science and Technology,  Skolkovo, Moscow Region, Russia
\AND
\Name{Svetlana Sushchinskaya} \Email{svetlana.sushchinskaya@skolkovotech.ru}\\
\addr Skolkovo Institute of Science and Technology,  Skolkovo, Moscow Region, Russia
\AND
\Name{Maxim Sharaev} \Email{m.sharaev@skoltech.ru}\\
\addr Skolkovo Institute of Science and Technology,  Skolkovo, Moscow Region, Russia
\AND
\Name{Alexander Andreev} \Email{aandreev@deloitte.ru}\\
\addr Deloitte Analytics Institute, Moscow, Russia
\AND
\Name{Alexey Artemov} \Email{a.artemov@skoltech.ru}\\
\addr Skolkovo Institute of Science and Technology,  Skolkovo, Moscow Region, Russia
\AND
\Name{Renat Akzhigitov} \Email{barms@yandex.ru}\\
\addr Moscow Research and Clinical Center for Neuropsychiatry, Moscow, Russia\\
Skolkovo Institute of Science and Technology,  Skolkovo, Moscow Region, Russia}
\editors{Alex Gammerman, Vladimir Vovk, Zhiyuan Luo, and Harris Papadopoulos}

\begin{document}

\maketitle
\medskip

\begin{abstract}
We consider a problem of diagnostic pattern recognition/classification from neuroimaging data. We propose a common data analysis pipeline for neuroimaging-based diagnostic classification problems using various ML algorithms and processing toolboxes for brain imaging. We illustrate the pipeline application by discovering new biomarkers for diagnostics of epilepsy and depression based on clinical and MRI/fMRI data for patients and healthy volunteers.
\end{abstract}


\section{Introduction} 
\label{sec:introduction}

The human brain is a complex system of interconnected and specialized structures, the functioning of which is associated with the numerous ongoing biophysical and biochemical processes. These processes differ in healthy people and in patients with various pathologies. Nowadays, the normal and pathological processes related to the brain structure and functioning could be recognized by analyzing the results of medical examination with the use of in-vivo scanning devices. 

In clinical practice, neuroimaging data of each patient is considered individually, either visually by doctor/neuroradiologist or by analyzing the clinically meaningful features (cortical volumes, thicknesses, etc.). Nowadays, Artificial Intelligence (AI), Machine Learning (ML) and Intelligent data analysis techniques are used in medical research for diagnostic biomarkers discovery and the treatment outcomes prediction with the use of neuroimaging data collected for the targeted groups of patients or healthy volunteers \cite{NeuroML1}.

In this article we confine ourselves to the problems of diagnostic pattern recognition/classification from neuroimaging data. The features (called biomarkers), which distinguish different groups of examined subjects, are extracted from neuroimaging data and further used in clinical practice for the diagnostic purposes. Biomarkers (characteristics which are objectively measured and evaluated as an indicator of normal biologic processes, pathogenic processes, or pharmacologic responses to a therapeutic intervention) are key components of modern medicine \cite{Bio2}. There is an ever-growing number of ML studies for detecting new clinically meaningful biomarkers from large neuroimaging datasets \cite{NeuroML1}.

Important key feature of neuroimaging data is its high dimensionality. For example, MRI signals for usual human brain with a volume of approximately $1.4\times10^6$ cubic millimeters are represented by a 3D array with a total dimensionality of the order of $10^7$, and fMRI images are represented by 4D-array of 3D images of lower resolution (about $10^5$ voxels) with a total dimensionality of the order of $10^7$. Thus, the curse of dimensionality phenomenon is often an obstacle for using ML techniques. To avoid this phenomenon, various universal dimensionality reduction methods \cite{DR1,DR2,DR3,chernova} and/or specific neuroimaging-oriented feature selection methods \cite{katya2} are used for extracting low-dimensional features from high-dimensional neuroimaging data. After that, ML algorithms are applied to these features. Such clinically meaningful features can be computed by brain image processing toolboxes \cite{MDA1,MDA2}.

As a result of ML application we obtain not only a classifier to support medical diagnosis, but also after posterior analysis of the classifier properties we identify features, which have biomedical interpretation and can be used for medical conclusions. Therefore, ML-based neuroimaging data processing for medical diagnostics  is a multistage iterative process, which uses various ML feature selection, extraction and classification algorithms, as well as domain-specific knowledge.

In this paper we propose a common data analysis pipeline for neuroimaging-based diagnostic classification problems using various ML algorithms and brain imaging toolboxes. The proposed pipeline consists of several stages, each of these stages can be executed several times in an iterative mode. Each stage, in turn, contains a number of different algorithms which can be split into levels, and each level consists of algorithms solving the same ML problem but based on different mathematical approaches. A composite algorithm, which performs (sequentially or in an iterative way) a data processing task using a particular set of algorithms from various stages and levels is called an algorithmic chain.

We illustrate the pipeline application by discovering new biomarkers for diagnostic of epilepsy and depression based on clinical and MRI/fMRI data for patients and healthy volunteers. This study was performed in collaboration with physicians from Russian Scientific and practical psycho-neurological center named after Z.P. Solovyov (NPCPN http://npcpn.ru/) (Skoltech biomedical partner), which provided medical data and biomedical expertise.

The paper is organized as follows. Section \ref{sec1} briefly describes main properties of processed MRI/fMRI data and provides some details of the medical diagnostic tasks; solutions of these tasks are considered later as examples. Section \ref{sec2} describes the main structure of the proposed pipeline (stages and levels of each stage) and specifies tasks solved at various levels. Section \ref{sec3} describes the first preprocessing stage in which various data cleaning procedures, neuroimaging toolboxes and dimensionality reduction procedures are used to obtain domain-specific features, which are used as inputs for further Exploratory ML/data analysis step. At this step, described in Section \ref{sec4}, various ML/data analysis techniques \cite{ML1,ML2,ML3,ML4,DR1,DR2,DR3} are applied to the neuroimaging data or   features extracted from it to select important features providing high classification accuracy. In Section \ref{sec5} conclusions are provided.

\section{Processed MRI/fMRI data}
\label{sec1}
\subsection{Properties of processed MRI/fMRI data}

Processed data consists of structural and functional MR images. The structural MRI protocols are aimed at providing the information concerning brain structure, enhancing required tissue patterns when using  different acquisition modalities. Functional MRI (fMRI) scanning regime is based on measurement of the blood oxygen-level dependent contrast (BOLD) that is directly related to neuronal activity and thus reflects brain functioning.

MRI data once acquired should be cleaned to eliminate the noise associated with the scanning procedure (low-level hardware artefacts such as magnetic field inhomogeneity, radiofrequency noise, surface coil artefacts and others) and signal processing (chemical shift , partial volume, etc.); besides there are artefacts associated with the scanned patient (physiological noise such as blood flow, movements, etc.) \cite{Erasmus2004}.  MRI/fMRI signals cleaning is one of the tasks solved at Preprocessing stage (see Section \ref{sec2} below). 

In addition to MRI data cleaning problem, there is another common challenge of the brain imaging analysis related to big data dimensionality, which mostly depends on resolution parameters of the scanner inductive detection coil. For instance, standard voxel sizes are within $0.5-2$ $mm^3$ in case of structural imaging (resulting in $10^7$ voxels for the whole brain volume) and $2-6$ $mm^3$ in a functional MRI series (resulting in $10^5$ voxels). Thus an MRI image, composed of huge number of small sized voxels, has higher spatial resolution and, hence, high dimensionality. To avoid the curse of the dimensionality phenomenon, ML methods are usually applied to lower dimensional features extracted from original scans by feature selection procedures. These procedures are also included into the Preprocessing stage.

\subsection{Data used in illustrative examples}

Data, provided by Skoltech’s biomedical partner (the NPCPN), consists of structural and functional MR images. The considered dataset contains structural MRI and resting state functional MRI images of $100$ patients: $25$ healthy volunteers and 25 patients with major depressive disorder in an acute depressive episode, as well as 25 epilepsy patients and 25 epilepsy patients with major depressive disorder. The dataset is enriched with clinical information including gender, age, disease duration, BDI (Beck Depression Inventory scaling) and other typical medical indicators. There are some patients with temporal lobe epilepsy (TLE) with and without MRI evidence for structural lesion (named TLE MRI Positive/ TLE MRI Negative groups)

In order to find functional biomarkers of depression and epilepsy we explored functional $T2^*$ MRI EPI series ($[64; 64; 30]$ voxels) repeated $133$ times with repetition time (TR) of $3.7$ seconds, and $T1$ weighted MPR images ($[352; 384; 384]$ voxels). The structural data was preprocessed in FreeSurfer toolbox \cite{FSL2}, resulting in a vector of morphological features with dimensionality $1\times894$. The functional data was preprocessed in Nilearn toolbox \cite{nilearn}, and functional connectivity graph based features were retrieved using Networkx library \cite{networkx} resulting in a vector of dimension $1\times 587$.

We considered different diagnostic tasks, taking for each of them the data from different subgroups of patients as inputs to ML procedures:
\begin{itemize}
\item Epilepsy (including patients with Depression) versus No Epilepsy (including patients with Depression) classification (EvsNE $50/50$ subjects);
\item Depression (including patients with Epilepsy) versus No Depression (including patients with Epilepsy) classification (DvsND $50/50$ subjects);
\item Epilepsy versus Healthy Control classification (EvsH $25/25$ subjects);
\item Depression versus Healthy Control classification (DvsH $25/25$ subjects);
\item Temporal Lobe Epilepsy (including patients with Depression) versus Healthy Control classification (TLEvsH $30/25$ subjects);
\item MRI Positive Temporal Lobe Epilepsy (including patients with Depression) versus Healthy Control classification (TLEPvsH $16/25$ subjects);
\item MRI Negative Temporal Lobe Epilepsy (including patients with Depression) versus Healthy Control classification (TLENvsH $64/25$ subjects);
\item Non Temporal Lobe Epilepsy (including patients with Depression) versus Healthy Control classification (NTLEvsH $20/25$ subjects);
\end{itemize}
and others.

\section{Pipeline structure for neuroimaging-based Machine Learning diagnostics}
\label{sec2}
Neuroimaging-based Machine Learning diagnostic task formulation consists of the following elements:
\begin{itemize}
\item List of possible diagnostic inferences (diagnoses, hypotheses, etc.), which are tested for an individual/patient by analyzing his/her neuroimaging and clinical data. For example, a person has a specific disease such as depression (D) or epilepsy (E) vs he/she is healthy (H) (abbreviations DvsH and EvsH, respectively); or the presence or absence of depression in epilepsy patients (abbreviation DEvsE),
\item Dataset with appropriate neuroimaging and clinical data, collected for target groups of subjects with known diagnostic inferences from the considered list (in case of examples, given above, this should be a group of patients with depression or epilepsy and a group of healthy volunteers; a group of patients with established depression and epilepsy diagnosis and a group of patients with epilepsy but without depression).
\end{itemize}
The task is to establish a diagnosis for previously unseen patient from his/her neuroimaging and clinical data. In Machine learning terms, this task is reduced to a supervised classification problem based on labeled data.

Proposed pipeline for the solution of this task consists of four stages, namely:
\begin{itemize}
\item Preprocessing stage;
\item Exploratory Machine Learning/Data analysis stage;
\item Inference stage;
\item Quality assessment stage.
\end{itemize}

Each stage, in turn, contains a number of different algorithms which can be split into levels, and each level consists of algorithms solving the same ML problem but based on different mathematical approaches.

At the Preprocessing stage MRI/fMRI data is transformed into various representations which then will be used as inputs for chosen Machine Learning procedures. This stage has several goals:
\begin{itemize}
\item neuroimaging data cleaning (denoising/noise reduction, removal of artefacts)using brain imaging software \cite{FSL2, katya3,katya4,katya5,katya6,MDA1,MDA2} and other data analysis techniques \cite{ArtemovSharaev};
\item  transforming original high-dimensional data into biomedically motivated brain characteristics (clinically meaningful features) with lower dimensionality such as vectors consisting of volumetric characteristics of chosen brain areas (Hippocampus, Lateralorbitofrontal, etc.), connectivity matrices, directed graph describing the brain connectome and preserving directions of information transfer also using brain imaging software \cite{MDA1,MDA2}. We call such features a priori domain-specific features;
\item computing new mathematical characteristics of the constructed mathematical objects (vectors, matrices, graphs) which describe various clinically meaningful properties of these objects (for example, constructing directed flag complex from the directed graphs representing connectivity among brain areas and computing its persistent homology characteristics such as Betti numbers, Euler characteristic, etc. \cite{B1}). These topological features  are now used in neuroimaging studies for discovering ``deep'' structure of the brain connectomes and are thought to be promising diagnostic biomarkers \cite{PTM1,PTM2,PTM3,PTM4}.
\item transforming original high-dimensional data or clinically meaningful features into their low-dimensional representations to avoid the curse of the dimensionality, by preserving clinically meaningful information using various feature selection/dimensionality reduction techniques \cite{DR1,DR2,DR3, katya2, MLM5,MLM6}.
\end{itemize}

The result of this stage are datasets consisting of objects such as vectors, matrices, graphs, with common name ``Machine Learning Input'' (MLI) data. The details of this stage are given in Section \ref{sec3} below.

In Exploratory Machine Learning/Data analysis stage, given constructed MLI-datasets, various ML techniques are applied to them. Obviously, the choise of the algorithm depends on the data structure: Support Vector Machine Classifier (SVC) \cite{SVC}, Logistic Regression Classifier (LR) \cite{LR}, Random Forest Classifier (RFC) \cite{RFC}, K Nearest Neighbors Classifier (KNN) \cite{KNN}, Extra Trees Classifier (ETC) \cite{ETC}, Neural Networks \cite{annInitialization,Ensembles} including 3D Deep convolutional neural networks \cite{3DDL} as well as anomaly detection and imbalanced classification methods \cite{oneclassP,modelselection,resampling} are applied to vectorized MLI-data; kernel-based classifiers \cite{K-BC}, \cite{Wang} are applied to connectivity graphs with different graph kernels \cite{GK}.

Each of the performed Machine Learning experiments is defined by a chosen triplet (Classification task, MLI-datasets, Machine Learning algorithm). For example, the triplet (EvsH, DS name, RFC) means that dataset with specific name (see Section \ref{sec3} for details) is used for establishing diagnosis Epilepsy using Random Forest Classifier algorithm. 

Each of the used algorithms is defined by a number of ``free parameters'' and their ``optimal'' values  are determined during experiments. Common multiple-fold cross-validation technique \cite{multiple-foldCrossValid} is usually used for this purpose.

The result of performed Machine Learning experiment is the constructed classifier (with tuned parameters) and its quality characteristics estimated using  final cross-validation procedure (for example, leave-one-out cross-validation \cite{Leave-one-outCrossValid}). If possible, we also extract clinically meaningful features (called a posteriori task-specific features), based on which the classifier makes its decision. 

In addition, subject-oriented classification results are collected from all performed experiments and are saved in specific Personal Classification Quality (PCQ) table. The table includes anonymized information about all subjects whose neuroimaging and clinical data is used in the study. Each row of this table corresponds to a particular subject,  columns of the table are split into groups. Zero group contains personal information about individuals (their personal identifiers, $ID$'s), and, if it is convenient for subsequent analysis of the table, a part of their clinical data --- for example, clinical status CSID ($E$, $D$, $H$, $DE$, or other), considered as labels in classification tasks. Each other group corresponds to results of a particular performed ML experiment ($MLE$). A sub-table, defined by $(ID, MLE)$, contains the following elements: a symbol $IID,MLE$ indicating whether the data of the subject ID is used in the experiment ($IID,MLE = 1$) or not ($IID,MLE = 0$); the number of cross-validation (CV) experiments $NCV,ID,MLE$ in which subject $ID$ “participated”; the numbers $NCV,ID,MLE,CSk$ of cross-validation experiments (among $NCV,ID,MLE$) where a particular classifier makes decision $CSk$, $k = 1, 2$.

Statistical analysis of this table allows making various conclusions. For example, let us assume that the clinical status $CS2$ is $H$ (healthy). Then averaged frequencies
\begin{equation}
\label{eq1}
P_{MLE,CS/CSk} = \frac{\sum_{ID:I_{ID,MLE} = 1,CS_{ID} = CS}N_{CV,ID,MLE,CSk}}{\sum_{ID:I_{ID,MLE} = 1,CS_{ID} = CS}N_{CV,ID,MLE,CSk}},
\end{equation}
where $CS$ takes values $CS1$ and $CS2 = H$, are equal to True Positive (TP) and False Positive (FP) rates of the used classifier when  $CS_{ID} = CS1$ and $CS_{ID} = CS2$, respectively.

Also for the considered classifier we can perform statistical analysis of the set $S_{MLE,CS1}$, consisting of frequencies 
\begin{equation}
\label{eq2}
P_{MLE,IDJ,CS1} = \frac{N_{CV,IDj,MLE,CS1}}{N_{CV,IDj,MLE}},\, j = 1,2,\ldots, N_{MLE,CS1},
\end{equation}
of TP decisions for individuals with clinical status $CS1$, whose data is used in the experiment MLE. Here $N_{MLE,CS1} = \sum_{ID:I_{ID,MLE} = 1,CS_{ID} = CS1}I_{ID,MLE}$ is the number of such individuals. These characteristics make it possible to understand if the classifier works ``statistically equally'' for all such individuals or not.

If based on \eqref{eq2} we make a positive conclusion about statistically comparable results, then we can use the dataset $S_{MLE,CS1}$ to estimate accuracy of the computed TP and FP rates \eqref{eq1} for the constructed classifier, as well as construct prediction regions for these rates for new individuals using conformal prediction framework \cite{Conf1,Conf2,Conf3,vovk,conformalized}.

If based on \eqref{eq2} we make a negative conclusion, then the individuals can be split into clusters with ``approximately equal'' personal qualities of classification. In the inference stage this allows
\begin{itemize}
\item to find possible dependencies between personal quality of classification (for a considered classifier) of the individual and his/her clinical data,
\item to construct ensembles of classifiers (if the clusters for different classifiers differ between themselves) using personal clinical data of an individual as additional input parameters when the ensemble is applied to calculate predictions for this individual. 
\end{itemize}

The details of this stage are given in Section \ref{sec4}.

In the Inference stage, given a number of constructed classifiers, we discover a posteriori task-specific clinically meaningful features (determined by specific classifiers), and Personal Classification Quality (PCQ) table containing results of all performed MLE. 

In this stage, final composite classifiers for a specific classification task are constructed using 
\begin{itemize}
\item either known Machine Learning approaches (e.g. ensembles of selected ``good'' classifiers, constructed for the same task in the previous stage with taking into account results of statistical analysis of the PCQ-table),
\item or by performing new MLE with input features, selected among discovered task-specific features, for both the considered task and other ``clinically related'' tasks.
\end{itemize}

For example, in case of the EvsDE classification task (diagnostics of depression for patients with epilepsy), task-specific features (including features from ``clinically related'' classification tasks) can be selected among features of
\begin{itemize}
\item EvsDE classification task (based on MRI-data),
\item EvsH classification task (based on MRI-data),
\item DvsH classification task (based on fMRI-data),
\end{itemize}
and can be used as a set of new ``combined'' features to improve solution of the EvsDE classification task. 

Classification results, corresponding to each particular subject, are saved in the PCQ table and can be used to estimate classification quality and to choose the most accurate classifier.

Note that usage of the same data in multiple successively executed steps can lead to over-fitting and, therefore requires both stratified division of samples into training/validation and testing sub-samples in cross-validation procedures and multiple cross-checks. 

The MLE performed with such combined features showed this approach to be promising in discovering neuroimaging-based biomarkers in neurology and psychiatry, see details in Section \ref{sec4}.



\section{Preprocessing stage}
\label{sec3}

Preprocessing stage has two main goals:  MRI/fMRI data cleaning  and avoiding the curse of the dimensionality phenomenon caused by high dimensionality of initial MRI/fMRI data. The latter goal can be achieved by constructing lower dimensional biomedically significant brain characteristics from the initial data.

\subsection{MRI/fMRI data cleaning}

Data cleaning include procedures for  3D MRI images denoising and removing artefacts (caused by various reasons) from 4D fMRI signals.

\textbf{MRI data cleaning.} An artefact is a feature appearing in an image that is not present in the original object. Depending on their origin, artefacts are typically classified as patient-related (motion, blood flow), signal processing dependent (chemical shift, partial volume) and hardware-related (magnetic field inhomogeneity, radiofrequency noise, surface coil artefacts and others) \cite{Erasmus2004}.

Some denoising procedures are performed in MRI scanner using specialized software, installed on the scanner, e.g. see Siemens BLADE and 3D PACE procedures \cite{Hirokawa2008,Pipe1999}. These methods could slightly vary from one manufacturer to another and in different software versions. Among them there are approaches to motion correction \cite{Pipe1999}, field inhomogeneity correction \cite{Simmons1994,Vovk2007} and phase error correction \cite{Sven2013}. 

The obtained images could be further preprocessed using neuroimaging software (\cite{FSL2}, \cite{katya3}, \cite{katya4}, and other brain imagery processing software toolboxes \cite{MDA1,MDA2}) in order to perform other types of correction \cite{Klein2009}, increase signal-to-noise ratio and exclude data artefacts, see, for example, \cite{Salimi-Khorshidi2014,Sladky2011,Thirion2006}.

\textbf{fMRI data cleaning.} fMRI data is represented as a sequence of $T2^*$ weighted (see Section \ref{sec1}) images with lower than structural MRI spatial resolution, usually sampled every $2-3$ seconds. These images should also be preprocessed in order to exclude different sources of noise/artefacts both in scanner during acquisition to remove low-level hardware artefacts and after scanning in neuroimaging software (\cite{katya3}, \cite{katya4}, \cite{katya5}, \cite{katya6} and other brain imaging toolboxes \cite{MDA1,MDA2}).

Initial fMRI data has complex multidimensional spatiotemporal structure and consists of recorded multidimensional time series, each component of which characterizes brain activity associated with blood flow (hemodynamic response) related to energy consumption by active cell clusters at specific brain voxel \cite{Huettel2004}. These measurements contain not only brain activity but also noise caused by various artefacts such as physiological (cardiac and respiratory) and non-physiological (movement, scanning artefacts, etc.) sources. In order to remove noise Independent Component Analysis (ICA)  \cite{ICA1} is used; extracted components are ordered according to the amount of explained variance; many of the first components will  not contain signal of interest. Often there is more noise components than signal, for example, $90\%$ of noise components \cite{Smith2013} and $88\%$ of noise components \cite{Griffanti2014} for multiband sequences, for extended discussion see \cite{ArtemovSharaev}.

\subsection{Constructing the subject-oriented (a priori domain-specific) features}

The goals of this sub-stage is to extract informative features (biomedically significant brain characteristics, clinically meaningful features) with lower dimensionality. The approach is typically realized in several steps:
\begin{itemize}
\item selection of an appropriate brain atlas \cite{A1,A2,A3,A4,A5,A6} which splits the brain into the anatomical areas  (e.g. Hippocampi, cortical areas and etc.), 
\item 3D MRI/4D fMRI images segmentation into disjoint sets (sub-images), consisting of voxels, corresponding to different brain regions (Regions of Interest, ROIs),
\item various characteristics calculation for each ROI or interaction (connectivity) between ROIs
\end{itemize}

Examples of such characteristics:
\begin{itemize}
\item	structural morphometric parameters (volumes, thicknesses, curvatures) of the selected anatomical areas from the MRI-image, which together form a volumetric vector. For example, MRI processing toolbox \cite{FSL2} parcels MRI images into regions corresponding to the chosen Desikan-Killiany atlas; calculates $7$ volumetric characteristics for each cortical region (NVoxels, Volume\_mm3, normMean, normStdDev, normMin, normMax, normRange) and $9$ geometric characteristics of subcortical regions (NumVert, SurfArea, GrayVol, ThickAvg, ThickStd, MeanCurv, GausCurv, FoldInd, CurvInd);
\item	functional connectivity parameters (see \cite{conn,nilearn}), which describe interactions between various functional areas and are based on various measures of dependency like Pearson correlation (or spectral coherence,  mutual information) between time series of fMRI signals,  which measure brain activity in chosen voxels from considered areas obtained from resting-state fMRI. These parameters are described by symmetric functional connectivity matrices (or undirected graphs). Functional connectivity graphs are then analyzed with special python software libraries such as \cite{networkx}. Thus, functional connectivity of each ROI could be represented via several basic graph features (clustering coefficient, local/global efficiency, degree/ closeness/betweenness centrality, average neighbor degree, etc.);
\item effective connectivity parameters (under causation concept) describing ``the influence one neural system exerts over another either directly or indirectly'' \cite{Friston2003}. This explicitly means that all links between brain areas have some direction, thus the brain connectome could be considered as a directed graph representing connectivity among neurons within the network and the information about the direction of information transfer is preserved. Methods of assessing effective connectivity are being developed nowadays, among them are model-based approaches, like dynamic causal modelling (DCM) \cite{Friston2003,Sharaev2016,Ushakov2016} and model-free approaches based on information theory \cite{Montalto2014,Sharaev2016m,Sharaev2018}.
\end{itemize}

For constructed objects (brain areas, symmetric connectivity matrices/undirected graphs, causal directed graphs) different characteristics reflecting meaningful properties of these objects, can be computed for further use in Machine learning studies:
\begin{itemize}
\item segments of MRI-image consisting of 3D MRI-voxels from chosen brain areas (to be used as inputs for deep learning procedures \cite{DL1,DL2,DL3});
\item various vector characteristics of undirected graphs (see \cite{Graph1}) with components describing various graph properties such as global/local node efficiency, cost, betweenness centrality, etc. \cite{Wang};
\item vectors consisting of persistent homology characteristics (such as Betti numbers, Euler characteristics, etc.) of directed flag complex \cite{PT1,PT2,PT3}, which are computed from the directed connectivity graphs (such characteristics are used in analysis of brain connectomes \cite{PTM1,PTM2,PTM3,PTM4}).
\end{itemize}

Most often domain-specific lower dimensional features (morphometric or functional connectivity features) could be extracted from original data in specialized MRI processing toolboxes \cite{MDA1,MDA2}.

\subsection{Low-dimensional representations of domain-specific features}

Although the dimensionality of constructed domain-specific features and corresponding characteristics can be low  in comparison with the initial data, it can nevertheless be rather high. For example, volumetric vector, computed by toolbox \cite{FSL2}, provides $897$ components. The size of connectivity matrix, computed by toolboxes \cite{conn,nilearn} is $116\times116$ (in accordance with the chosen brain atlas); for each node $6$ graph characteristics (measures of nodes centrality, local efficiency and others) and two ``global'' graph characteristics (characteristic path length and global efficiency) are computed producing a vector of dimensionality $698 = 116\times6+2$ \cite{sveta1}.

Luckily, these data, as well as the most real-world high-dimensional data obtained from ``natural'' sources (including MRI and fMRI data), due to dependencies between its components and various constraints on their values, do not fill the whole full-dimensional space and occupies only a very small domain  with smaller intrinsic dimension. Thus, such high-dimensional data can be transformed into some lower-dimensional representations (or features) using various Feature extraction/Dimensionality reduction algorithms \cite{DR2,DR3}.

If the data is concentrated near a linear low-dimensional affine subspaces, various linear methods can be used such as Principal Component Analysis (PCA) \cite{PCA}, Independent Component Analysis (ICA) \cite{ICA1}, Projection Pursuit \cite{PP1}, etc. But in many cases the ``low-dimensional area'' is essentially nonlinear and requires using advanced nonlinear Feature extraction/Dimensionality reduction algorithms. The most popular model of high-dimensional data, which occupy a small part of observation space, is a Manifold model in accordance with which the data is located  near an unknown Data manifold of lower dimension, embedded in an ambient high-dimensional input space \cite{ML1}; this manifold model can effectively represent the brain anatomy as well \cite{ML2}. Dimensionality reduction algorithms under this model, called Manifold learning \cite{ML4} are widely used for medical data preprocessing including MRI/fMRI data \cite{MLM2,MLM9}.

\section{Machine Learning/data analysis pipeline}
\label{sec4}

\begin{figure}
\centering
\includegraphics[width=0.5\textwidth]{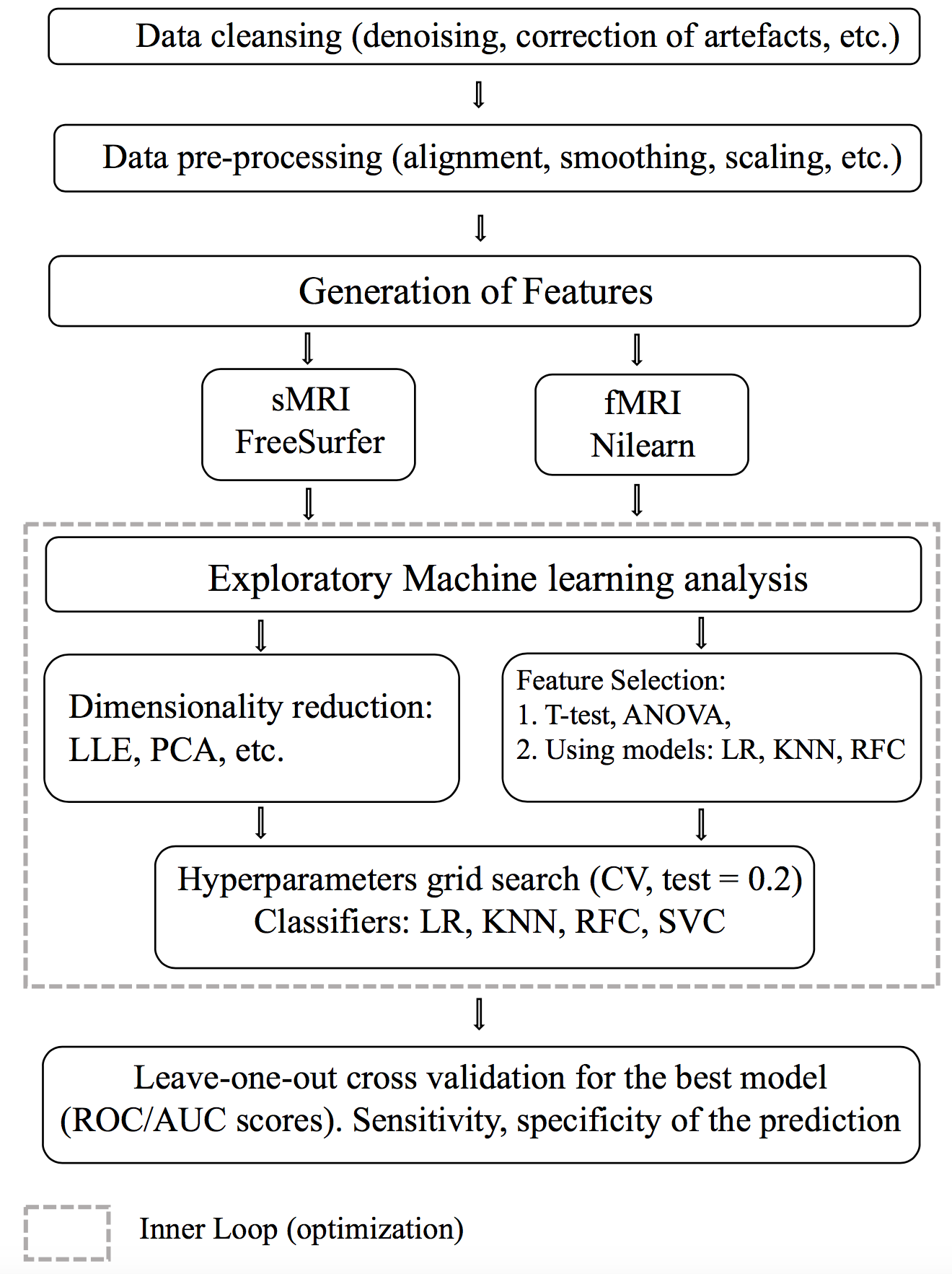}
\caption{\label{fig1} Flow diagram illustrating the classification pipeline}
\end{figure}

\begin{figure}
\centering
\includegraphics[width=0.5\textwidth]{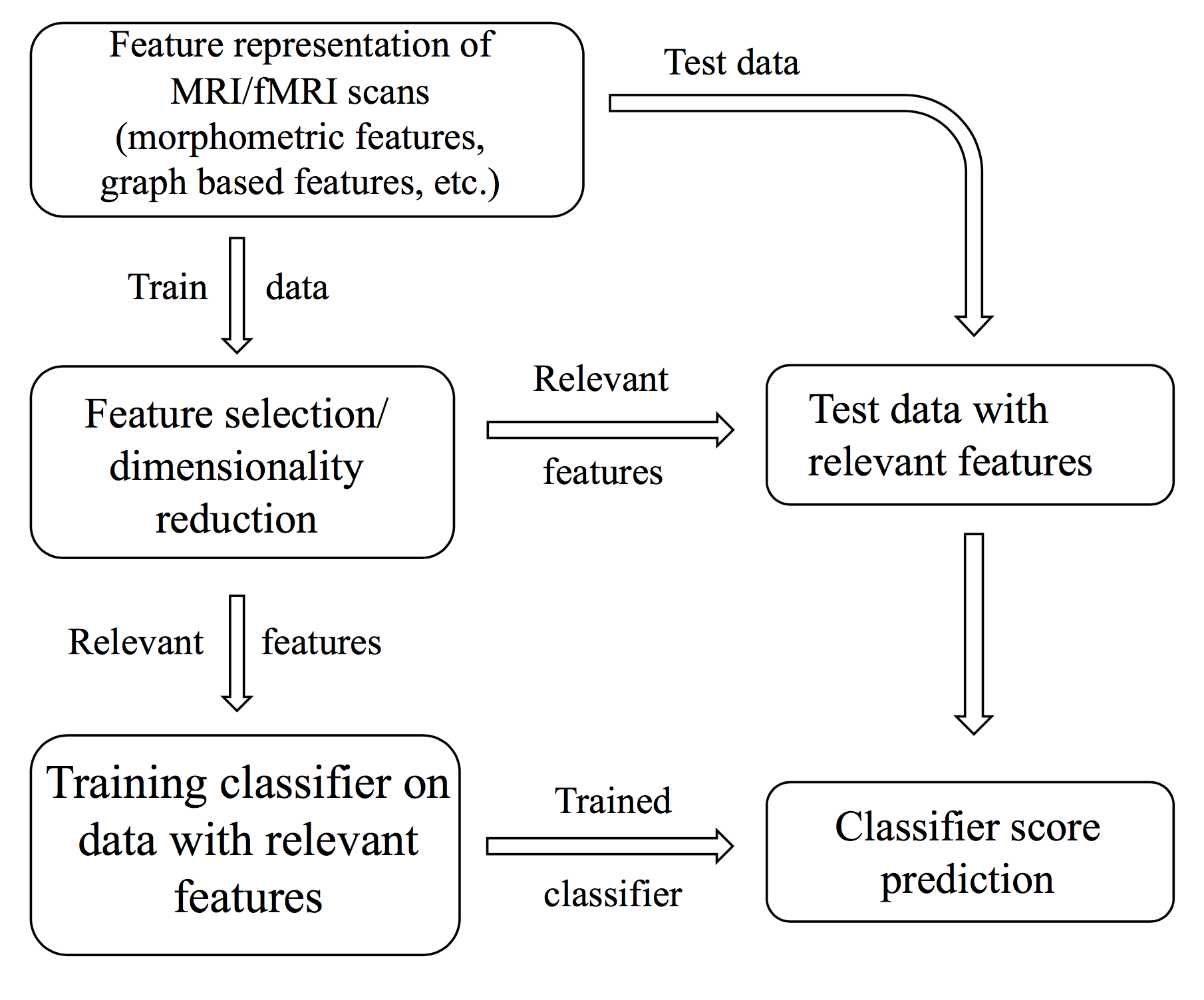}
\caption{\label{fig2} Inner loop recursive diagram illustrating the classifier hyper-parameters grid search}
\end{figure}

According to the proposed pipeline schema (see figure \ref{fig1} for details), the MRI data were cleaned, preprocessed and their features were extracted using MRI processing toolboxes \cite{MDA1,MDA2}.

Structural morphometric features were calculated from $T1w$ images using \cite{FSL2}; for more than $100$ brain regions corresponding features explaining brain structure (volumes, surface areas, thicknesses, etc.) were computed producing a vector with $894$ features for each subject.

Functional connectivity matrices were calculated from $T2^*$ EPI MRI sequences using \cite{nilearn} toolbox; functional connectivity matrix was considered as a graph with nodes in the corresponding regions of interest (ROI); for each node basic graph features (local/global efficiency, betweenness centrality, etc.) were calculated, thus producing a vector with $587$ features for each subject.

These datasets were investigated separately in order to evaluate informative content of each dataset. The Machine learning exploratory pipeline was realized in IPython using Sklearn library (\url{http://scikit-learn.org/}) and organized as follows (see figure \ref{fig2} for details): 
\begin{itemize}
\item We considered two geometrical methods for dimensionality reduction: 1) Locally Linear embedding; 2) Principal Component Analysis.
\item We considered two methods of feature selection: 1) Feature selection with SelectKBest() function, based on Pearson's chi-squared test and ANOVA scoring; 2) Selection of relevant features based on a particular classification model via the Sklearn function SelectFromModel(), used with Logistic Regression (LR), K-Nearest Neighbors (KNN) and Random Forest Classifier (RFC).
\item We performed grid search for a number of selected features in the set  $\{10, 20, 50, 100\}$
and for a number of components in dimension reduction procedure in the set $\{5, 10, 15, 20\}$.
\end{itemize}

Data was whitened before  training. Feature reduction was performed without double-dipping \cite{katya2}, therefore training and testing datasets are separated before feature selection/dimensionality reduction. Hyper-parameters grid search was based on cross-validation with stratification, repeated $10$ times for each person being in test. 

\subsection{Classification of the Epilepsy}

In table \ref{tt1} we provide results for EvsH classification task. 

\begin{table}[t]
  \centering
  \begin{tabular}{cc|cc}
    \toprule
     \multicolumn{2}{c|}{Functional connectivity graph}
      & \multicolumn{2}{c}{Brain morphometry features
 (MRI)}\\
      \multicolumn{2}{c|}{based features (fMRI)}
      & \multicolumn{2}{c}{features
 (MRI)}
    \\ 
    \midrule
False Positive Rate & True Positive Rate & False Positive Rate & True Positive Rate\\
    \midrule
10\% & 16\% & 10\% & 35\% \\
15\% & 24\% & 15\% & 44\% \\ 
20\% & 36\% & 20\% & 55\% \\
30\% & 52\% & 30\% & 71\% \\
\bottomrule
  \end{tabular}
  \caption{\label{tt1} Classification of Epilepsy versus Healthy Control (EvsH, $25/25$ persons) based on the structural and functional MRI features}
\end{table}

We obtain the most accurate results when using MRI structural features. Now let us consider classification of patients with temporal lobe epilepsy (TLE) with and without magnetic resonance imaging (MRI) evidence for structural lesion (TLE MRI Positive/ TLE MRI Negative). In table \ref{tt2} we provide results for classification of MRI positive TLE versus Healthy Control (TLEvsHC, $30/25$ persons).

\begin{table}[t]
  \centering
  \begin{tabular}{cc|cc}
    \toprule
     \multicolumn{2}{c|}{Epilepsy versus Healthy Control}
      & \multicolumn{2}{c}{Temporal Lobe Epilepsy versus Healthy}\\
      \multicolumn{2}{c|}{(EvsH, 25/25 person)}
      & \multicolumn{2}{c}{Control (TLEvsHC, 30/25 persons)}
    \\ 
    \midrule
False Positive Rate & True Positive Rate & False Positive Rate & True Positive Rate\\
    \midrule
10\% & 35\% & 10\% & 50\% \\
15\% & 44\% & 15\% & 57\% \\ 
20\% & 55\% & 20\% & 87\% \\
30\% & 71\% & 30\% & 87\% \\
\bottomrule
  \end{tabular}
  \caption{\label{tt2} Classification of MRI positive TLE versus Healthy Control using structural MRI}
\end{table}

In table \ref{tt3} we consider results of classification of MRI Negative TLE versus Healthy Control (TLENvsH, $14/25$ persons).

\begin{table}[t]
  \centering
  \begin{tabular}{cc|cc}
    \toprule
     \multicolumn{2}{c|}{MRI Positive TLE versus Healthy Control}
      & \multicolumn{2}{c}{MRI Negative TLE versus Healthy Control}\\
      \multicolumn{2}{c|}{(TLEPvsH, $16/25$ person)}
      & \multicolumn{2}{c}{(TLENvsH, $14/25$ person)}
    \\ 
    \midrule
False Positive Rate & True Positive Rate & False Positive Rate & True Positive Rate\\
    \midrule
10\% & 12\% & 10\% & 93\% \\
15\% & 50\% & 15\% & 93\% \\ 
20\% & 63\% & 20\% & 93\% \\
30\% & 75\% & 30\% & 93\% \\
\bottomrule
  \end{tabular}
  \caption{\label{tt3} Classification of MRI negative TLE versus Healthy Control using structural MRI}
\end{table}

Thus we can see that when dividing the TLE group into the positive and negative subsets we obtained that MRI Negative TLE classification shows considerably higher sensitivity and specificity then MRI Positive TLE. Thus, further investigation of extracted features can shed light on differences of subsets and explain these findings.

The most important features for MRI Positive classification are Right Cerebellum, Precuneus, Left Accumbens and Right Putamen.
The most important features for MRI Negative TLE are
Right and Left Amygdala, Frontal pole, Insula, Left Cerebellum, Parsorbitalis and Isthmus cingulate. Thus, the best classifier was constructed for TLE Negative Epilepsy classification. Its sensitivity is equal to $90\%$ and specificity is equal to $93\%$.

\subsection{Classification of Depression}
There are papers indicating that in case of depression special patterns in brain structure can be recognized \cite{dep1,dep2}. In table \ref{tt4} we provide results of depression classification using either structural MRI or functional fMRI data. We can see that fMRI data provides more informative biomarkers of depressive disorders.

\begin{table}[t]
  \centering
  \begin{tabular}{cc|cc}
    \toprule
     \multicolumn{2}{c|}{Brain morphometry}
      & \multicolumn{2}{c}{Functional connectivity graph}\\
      \multicolumn{2}{c|}{features (MRI)}
      & \multicolumn{2}{c}{based features (fMRI)}
    \\ 
    \midrule
False Positive Rate & True Positive Rate & False Positive Rate & True Positive Rate\\
    \midrule
10\% & 22\% & 10\% & 12\% \\
15\% & 32\% & 15\% & 40\% \\ 
20\% & 47\% & 20\% & 64\% \\
30\% & 65\% & 30\% & 80\% \\
\bottomrule
  \end{tabular}
  \caption{\label{tt4} Classification of Depression versus Healthy Control (DvsH, 25/25 person)}
\end{table}

The most important features for fMRI-based depression classification are Left Caudate, Left Temporal Pole, Right Insula and Right Superior Occipital gyrus. The best accuracy, achieved for depression classification, has $70\%$ sensitivity and $80\%$ specificity.

\section{Conclusions}
\label{sec5}

In this paper we proposed a data analysis pipeline for processing of MRI/fMRI data and diagnostic classification on its basis. We verified the pipeline by identifying biomarkers, relevant for detection of epilepsy and depression with sufficiently high accuracy. Further research direction will be to develop non-parametric algorithms for classification quality assessment (accuracy evaluation) based on conformal prediction framework and consider various topological features of MRI/fMRI data as biomarkers.

\acks\begingroup
This study was performed in the scope of the Project ``Machine Learning and Pattern Recognition for the development of diagnostic and clinical prognostic prediction tools in psychiatry, borderline mental disorders, and neurology'' (a part of the Skoltech Biomedical Initiative program). 
\endgroup


\bibliography{references/references.bib}

\end{document}